\documentclass[10pt]{article}

\usepackage[margin=0.72in]{geometry}
\usepackage[utf8]{inputenc}
\usepackage{microtype}
\usepackage{amsmath,amssymb,amsfonts}
\usepackage{bm}
\usepackage{graphicx}
\usepackage{xcolor}
\usepackage{tikz}
\usepackage{float}
\usepackage{booktabs}
\usepackage{tabularx}
\usepackage{array}
\usepackage{enumitem}
\usepackage[numbers,sort&compress]{natbib}
\usepackage{xurl}
\usepackage[colorlinks=true,linkcolor=blue,citecolor=blue,urlcolor=blue]{hyperref}
\usetikzlibrary{arrows.meta,positioning,fit,calc}

\newcolumntype{Y}{>{\raggedright\arraybackslash}X}
\newcolumntype{L}[1]{>{\raggedright\arraybackslash}p{#1}}
\newcommand{\scale}{SCALE}
\graphicspath{{figures_mh_v1/}{figures_v2/}}
\setlist{nosep,leftmargin=*}
\hypersetup{
  pdftitle={SCALE: Simulation-Calibrated Amortized Learning for Energy Materials},
  pdfauthor={Kuan Huang and Bo Bai},
  pdfsubject={Energy materials; simulation-calibrated amortized learning; transformer surrogate learning; metal hydride hydrogen storage},
  pdfkeywords={energy materials discovery, simulation-calibrated amortized learning, physics-to-transformer architecture, periodic graph transformer, operator-generated training data, transformer surrogate models, energy world model}
}

\title{\bfseries
\LARGE SCALE: Simulation-Calibrated Amortized\\
\LARGE Learning for Energy Materials\\[0.35em]
\large A hybrid architecture connecting deterministic modeling,\\
\large real-world data, and transformer-scale inference\\
\large for accelerated energy-materials discovery}

\author{
Kuan Huang$^{1*}$ \quad Bo Bai$^{1}$\\[0.4em]
\small $^1$Alpha Ladder Particle Pte. Ltd., 168 Robinson Road, \#26-02, Singapore 068912\\
\small $^*$Corresponding author: \texttt{khuang@alphaladderparticle.ai}
}

\date{September 3, 2026}

\begin{document}
\maketitle

\begin{abstract}
Energy systems face converging pressures for security, affordability, resilience, and sustainability. Achieving the energy transition requires faster discovery of materials that can be deployed at scale. Yet innovation is constrained by an evidence-computation imbalance: experiments provide decisive real-world evidence but are sparse, heterogeneous, and protocol-dependent; deterministic physics and chemistry models are mechanistically informative but expensive across large candidate spaces; and direct machine learning is fast but limited by label fidelity and coverage.

Here we introduce \textbf{\scale} (\textbf{S}imulation-\textbf{C}alibrated \textbf{A}mortized \textbf{L}earning for \textbf{E}nergy Materials), a physics-grounded, real-world-data-calibrated transformer architecture for accelerated energy-materials discovery. \scale{} connects executable deterministic operators, experimental calibration, expanded calibrated virtual labeling, and transformer distillation in an auditable framework for rapid inference, ranking, inverse design, and active learning. We formulate the architecture, identify ten method-based application regimes, and present a representative implementation for solid-state metal-hydride hydrogen-storage capacity prediction.

In this implementation, a hydride phase-equilibrium capacity operator is calibrated against 381 measured capacity anchors from ML-HydPARK. The calibrated operator populates 5,000 candidate-condition-prototype teacher labels. Rather than represent candidate chemistry as formula text, SCALE prepares crystallographically anchored parent-alloy periodic graphs that preserve atomic sites, periodic neighbor relationships, and local metal environments relevant to hydrogen accommodation, then applies an edge-biased graph transformer to this representation. This periodic-graph-transformer approach proved effective in the present implementation: a 2.90-million-parameter graph transformer reproduces calibrated teacher labels with five-fold surrogate fidelity of MAE $0.0582$ wt\% H$_2$, RMSE $0.0833$ wt\% H$_2$, $R^2=0.9927$, and Pearson $r=0.9963$. Post hoc attention analysis further suggests that, without chemistry-group labels as supervision, SCALE learns chemically organized element groupings and metal--metal relationships consistent with established hydride chemistry. Once trained, SCALE reduces million-candidate evaluation from repeated deterministic workflow execution to batched learned inference, lowering per-candidate screening cost by approximately $10^7$--$10^8$ while retaining explicit links to the underlying simulation and experimental evidence.
\end{abstract}


\clearpage
\twocolumn

\section{Introduction}

The global energy system confronts a tightening dual imperative: supply must expand rapidly to sustain electrification and AI-intensive growth, while simultaneously becoming more secure, resilient, affordable, and substantially less carbon-intensive. Energy security is under growing pressure from supply disruptions, geographically concentrated production and processing, and vulnerabilities in critical energy infrastructure \citep{iea2025weo}. Electricity demand is also being reshaped by artificial-intelligence data centers, which are emerging as exceptionally large and geographically concentrated loads: global data-center consumption rose by 17\% in 2025 and is projected to nearly double from approximately 485 TWh in 2025 to 950 TWh by 2030, with demand from AI-focused facilities roughly tripling over the same period \citep{iea2026energyai}. This demand growth is occurring while global energy-related carbon dioxide emissions remain at record levels, reaching 38.4 Gt in 2025 and underscoring the persistent need to decarbonize energy supply \citep{iea2026globalenergy}. Energy materials lie at the center of these converging pressures because their composition, structure, defects, and interfaces govern conversion efficiency, storage density, safety, degradation, resource intensity, and operating lifetime \citep{yao2023sustainableenergy,ku2024scarcity}. Electrode and electrolyte chemistries set the energy density, cost, thermal stability, and cycle life of batteries \citep{innocenti2024battery}; absorber and interfacial materials dictate the efficiency, stability, and manufacturability of emerging photovoltaics \citep{jiang2024perovskite}; and catalysts, membranes, and storage materials determine the efficiency and practicality of hydrogen production, conversion, transport, and storage \citep{broom2026hydrogen}. Accelerating the reliable discovery of energy materials is therefore a system-level imperative rather than an isolated materials-development objective.

Artificial intelligence for science has made materials discovery more programmable, searchable, and model-driven. High-throughput repositories and workflow engines now organize first-principles calculations at scale \citep{jain2013materialsproject,curtarolo2012aflow,kirklin2015oqmd,mathew2017atomate}; supervised and graph-based models accelerate property prediction across molecules, polymers, and crystals \citep{butler2018mlmaterials,schmidt2019mlsolid,ward2016generalpurpose}; learned interatomic potentials extend atomistic simulation beyond routine first-principles length and time scales \citep{chen2022m3gnet,deng2023chgnet,yang2024mattersim}; generative models propose new inorganic candidates \citep{merchant2023gnome,zeni2025mattergen}; and operator-learning methods, physics-informed networks, and active-learning loops provide mathematical machinery for surrogate modeling, inverse problems, and adaptive data acquisition \citep{raissi2019pinn,li2021fno,lu2021deeponet,settles2009active,frazier2018bo}. Yet reliable scientific AI cannot depend on statistical association alone; it must incorporate physical and chemical principles or remain auditable against them. Physics-informed neural networks pioneered the direct incorporation of governing equations into learning, while multi-fidelity and $\Delta$-machine-learning methods connect evidence across different levels of accuracy; however, these approaches are typically tailored to specific equations, domains, or tasks and do not yet constitute broadly reusable materials foundation models \citep{raissi2019pinn,kennedy2001calibration,ramakrishnan2015delta,perdikaris2017multifidelity,pilania2017multifidelity}. The development of broadly reusable scientific models remains constrained by the availability, quality, and coverage of experimentally grounded training data \citep{pyzerknapp2025foundation}. Experimental synthesis and characterization are resource-intensive and low-throughput, while high-fidelity deterministic calculations demand substantial high-performance-computing resources and become prohibitive when repeated across combinatorial spaces of structures, compositions, and operating conditions \citep{butler2018mlmaterials,schmidt2019mlsolid,yao2023sustainableenergy}. Overcoming these coupled evidentiary and computational constraints therefore requires a general methodology capable of integrating mechanistic calculations and experimental measurements into scalable, fidelity-calibrated training data, and transferring that evidence into advanced learning architectures, such as graph neural networks and multimodal foundation models, for rapid and transferable materials inference \citep{chen2022m3gnet,pyzerknapp2025foundation}.

In this paper, we present \textbf{\scale} (\textbf{S}imulation-\textbf{C}alibrated \textbf{A}mortized \textbf{L}earning for \textbf{E}nergy Materials), a calibrated physics-to-learning framework for energy-materials discovery. We first formulate the architecture as a general scientific-computing framework. We identify and define ten method-based benchmark regimes for its application. We then present a representative metal-hydride implementation combining a thermodynamic deterministic model, experimental calibration, calibrated dataset construction, and graph-transformer learning. From this implementation, we assess learning fidelity, scientific interpretability, computational and financial scaling, and broader validation requirements and limitations of the SCALE framework.

\section{SCALE Framework and Benchmark Regimes}

\scale{} is proposed as a hybrid architecture that connects mechanistic physical and chemical knowledge encoded in deterministic models, experimental evidence, and amortized learning to address the evidence sparsity, computational cost, and deployment-speed challenges outlined above. It treats deterministic scientific models as executable operators, experimental measurements as calibration anchors, calibrated operator outputs as scalable training evidence, and advanced learning models as low-cost inference engines. The aim is to convert selected high-cost scientific evidence into reusable models that can support rapid screening, ranking, inverse design, uncertainty triage, and active-learning decisions across energy-materials problems.

\subsection{\scale{} Architecture}

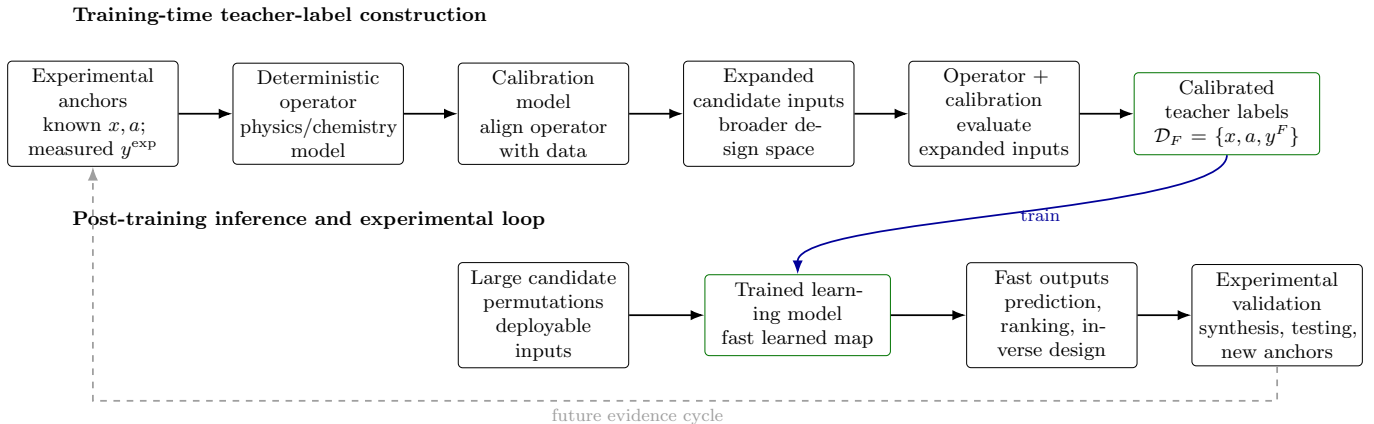
\begin{figure*}[t]
\centering
\resizebox{\textwidth}{!}{%
\begin{tikzpicture}[
  node distance=0.72cm and 0.86cm,
  box/.style={draw,rounded corners=2pt,align=center,minimum height=0.95cm,text width=2.48cm,font=\small},
  greenbox/.style={draw=green!45!black,rounded corners=2pt,align=center,minimum height=0.95cm,text width=2.72cm,font=\small},
  arrow/.style={-{Latex[length=2.2mm]},thick},
  softarrow/.style={-{Latex[length=2.2mm]},thick,dashed,gray!75}
]
\node[font=\bfseries\small,anchor=west] at (-0.45,1.55) {Training-time teacher-label construction};
\node[box] (anchors) {Experimental anchors\\known $x,a$; measured $y^{\exp}$};
\node[box,right=of anchors] (operator_a) {Deterministic operator\\physics/chemistry model};
\node[box,right=of operator_a] (cal) {Calibration model\\align operator with data};
\node[box,right=of cal] (expanded) {Expanded candidate inputs\\broader design space};
\node[box,right=of expanded] (operator_b) {Operator + calibration\\evaluate expanded inputs};
\node[greenbox,right=of operator_b] (labels) {Calibrated teacher labels\\$\mathcal{D}_{F}=\{x,a,y^F\}$};

\node[font=\bfseries\small,anchor=west] at (-0.45,-1.70) {Post-training inference and experimental loop};
\node[box,below=1.55cm of cal] (permutations) {Large candidate permutations\\deployable inputs};
\node[greenbox,right=1.20cm of permutations] (model) {Trained learning model\\fast learned map};
\node[box,right=1.20cm of model] (outputs) {Fast outputs\\prediction, ranking, inverse design};
\node[box,right=of outputs] (validation) {Experimental validation\\synthesis, testing, new anchors};

\draw[arrow] (anchors) -- (operator_a);
\draw[arrow] (operator_a) -- (cal);
\draw[arrow] (cal) -- (expanded);
\draw[arrow] (expanded) -- (operator_b);
\draw[arrow] (operator_b) -- (labels);
\draw[arrow] (permutations) -- (model);
\draw[arrow] (model) -- (outputs);
\draw[arrow] (outputs) -- (validation);
\draw[arrow,blue!60!black] (labels.south) .. controls +(0,-0.90) and +(0,0.90) .. node[right,font=\footnotesize,text=blue!60!black]{train} (model.north);
\draw[softarrow] (validation.south) |- ++(0,-0.55) -| node[pos=0.27,below,font=\footnotesize]{future evidence cycle} (anchors.south);
\end{tikzpicture}%
}
\caption{\scale{} separates calibrated evidence generation from deployment inference. Deterministic operators and experiments are used offline to construct calibrated teacher labels for training; after training, large candidate permutations are evaluated by the learned model, and active-learning outputs return to experimental validation.}
\label{fig:scale_flow}
\end{figure*}

The architecture of \scale{} is presented in Figure~\ref{fig:scale_flow}. A material candidate and its condition vector first enter a chemistry- or physics-grounded deterministic operator, which provides mechanistic predictions and structured physical states. Selected high-fidelity experimental measurements are then used to develop an additional calibration layer that corrects systematic deviation in the operator response. Together, the deterministic operator and calibration model establish a mechanistically grounded and experimentally anchored labeling process for the target material property, although executing this process repeatedly across large candidate spaces can remain computationally costly. To build a dataset large enough for training an advanced learning model, expanded candidate inputs are generated and passed through the deterministic operator and calibration layer to produce fidelity-calibrated teacher labels. Finally, a transformer-based model is trained to learn these teacher labels from deployable material and condition inputs, enabling rapid inference, ranking, inverse design, and active-learning selection without rerunning the deterministic workflow for every new candidate.

To formalize this workflow, let $x$ denote a material candidate and $a$ denote the operating, synthesis, processing, or device conditions under which the candidate is evaluated. The real response is
\begin{align}
  y^{\mathrm{real}} = U(x,a),
  \label{eq:real_response}
\end{align}
where the response may be a scalar property, a spectrum, a time series, a field, or a multi-objective decision vector.

\paragraph{A. Deterministic operator.}
A chemistry- or physics-grounded deterministic model maps candidate-condition inputs to structured operator outputs:
\begin{align}
  \bm{s}_i=\mathcal{S}(x_i,a_i;\alpha),
  \label{eq:det_operator}
\end{align}
where $\alpha$ denotes modeling choices such as force field, functional, mesh, timestep, convergence threshold, kinetic mechanism, or boundary condition. The output $\bm{s}_i$ can include the deterministic prediction, latent physical states, convergence diagnostics, uncertainty proxies, descriptors, and quality-control flags.

\paragraph{B. Experimental calibration model.}
Given an anchor set of paired deterministic outputs $\bm{s}_i$ and experimental measurements $y_i^{\exp}$, \scale{} evaluates a family of candidate calibration models and selects the best-performing calibrator to map structured operator outputs onto measured responses:
\begin{align}
  \phi^\star
  =
  \arg\min_{\phi \in \Phi}
  \sum_{i=1}^{n_{\exp}}
  \mathcal{L}\!\left[
    C_\phi(x_i,a_i,\bm{s}_i),
    y_i^{\exp}
  \right]
  +\Omega(\phi).
  \label{eq:calibrator}
\end{align}
Here $\Phi$ denotes the candidate calibrator family, $\mathcal{L}$ measures prediction error against experiment, and $\Omega$ penalizes unnecessary model complexity. $C_\phi$ may be statistical, tree-based, probabilistic, kernel-based, or neural. Its role is not to replace the deterministic model; its role is to learn systematic residuals between the modeled response and experimental reality.

\paragraph{C. Expanded teacher-label generation.}
After calibration, \scale{} constructs an expanded input set
\begin{align}
  \mathcal{X}_{\mathrm{virt}}=\{(x_j,a_j)\}_{j=1}^{N},
  \label{eq:virtual_input_set}
\end{align}
covering additional candidate structures, compositions, and operating conditions. Each expanded input is evaluated by the deterministic operator and then passed through the selected calibration model to produce a fidelity-calibrated teacher label:
\begin{align}
  y_j^{F}=C_{\phi^\star}\!\left(x_j,a_j,\mathcal{S}(x_j,a_j;\alpha)\right).
  \label{eq:calibrated_label}
\end{align}
Together, these input-label pairs form the fidelity-calibrated training corpus used for model training.

\paragraph{D. Transformer model and representation design.}
Before training, each candidate-condition input is encoded into a representation appropriate to the material class, structure, and target property:
\begin{align}
  \bm{h}_{j,1:m}=E_\psi(x_j,a_j).
  \label{eq:encoder}
\end{align}
Here $E_\psi$ converts the input into $m$ learnable tokens or structured embeddings. It may represent molecules as sequence or descriptor tokens, polymers as repeat-unit, composition, and condition tokens, crystals as periodic graphs, devices as coupled component graphs, or spatiotemporal systems as field or operator tokens. The encoder is not cosmetic preprocessing; it determines which physical and chemical relationships the learning model can access.

The transformer updates these token representations through $L$ layers:
\begin{align}
  \bm{Z}^{(0)}
  &=
  \bm{h}_{j,1:m},
  \nonumber\\
  \bm{Z}^{(\ell+1)}
  &=
  \mathrm{TransformerBlock}_{\ell}
  \!\left(\bm{Z}^{(\ell)}\right),
  \nonumber\\
  &\hspace{1.2em}\ell=0,\ldots,L-1.
  \label{eq:transformer_update}
\end{align}
The calibrated target is then predicted through a pooling operation and prediction head:
\begin{align}
  \hat{y}_j
  =
  g_\omega\!\left(\mathrm{Pool}(\bm{Z}^{(L)})\right).
  \label{eq:prediction_head}
\end{align}
Training selects the encoder, transformer, and prediction-head parameters by minimizing error against the fidelity-calibrated teacher labels:
\begin{align}
  \theta^\star,\psi^\star,\omega^\star
  =
  \arg\min_{\theta,\psi,\omega}
  \sum_{j=1}^{N}
  \mathcal{L}\!\left[
    \hat{y}_j,
    y_j^{F}
  \right].
  \label{eq:transformer_objective}
\end{align}

\paragraph{E. Amortized learning and large-scale inference.}
After training, \scale{} uses the learned model directly for large candidate permutations:
\begin{align}
  \hat{y}_{k}
  =
  g_{\omega^\star}\!\left(
  \mathrm{Pool}\!\left[
  F_{\theta^\star}\!\left(E_{\psi^\star}(x_k,a_k)\right)
  \right]\right),
  \nonumber\\
  k=1,\ldots,N_{\mathrm{screen}}.
  \label{eq:screening_inference}
\end{align}
This is the amortization step in \scale{}: the expensive cost of deterministic computation and experimental calibration is paid during teacher-label construction and model training, then reused across many low-cost inference calls. Because batched inference is much cheaper than rerunning the deterministic operator, the trained model can support rapid screening, ranking, inverse design, uncertainty triage, and selection of candidates for experimental validation. New measurements can then enter a later \scale{} cycle as additional calibration anchors.

\subsection{Benchmark Regimes}

The structure of \scale{} makes it portable across energy-materials development because the architecture is organized around the scientific operator, not a single material family or software package. Once a deterministic workflow can map a candidate and condition to a structured output, and once selected experiments can calibrate the relevant response, the calibrated workflow can be distilled into transformer-scale inference. This enables the same framework to support discovery problems in storage, hydrogen, carbon capture, solar conversion, thermal management, fusion-facing materials, and electrochemical devices. Table~\ref{tab:scale_categories} lists ten method-based benchmark regimes where \scale{} can be instantiated. The regimes are deliberately organized by modeling principle rather than by material class: a solid electrolyte, for example, may require electronic-structure calculation for phase stability, atomistic simulation for ion transport, continuum mechanics for stack contact, and device modeling for cell-level behavior.

\begin{table*}[t]
\centering
\caption{Method-based benchmark regimes for \scale. Compute-density values are order-of-magnitude planning estimates if only deterministic models are used directly for screening before transformer amortization; actual cost depends on system size, convergence criteria, hardware, licensing, and campaign design.}
\label{tab:scale_categories}
\tiny
\setlength{\tabcolsep}{1.6pt}
\renewcommand{\arraystretch}{1.15}
\begin{tabularx}{\textwidth}{L{0.125\textwidth}L{0.165\textwidth}L{0.215\textwidth}L{0.315\textwidth}L{0.140\textwidth}}
\toprule
Category & Representative deterministic tools/workflows & Representative energy materials & Representative studies and \scale{} role & Deterministic workload \\
\midrule
Electronic-structure screening & VASP, Quantum ESPRESSO, CP2K, CASTEP, ABINIT, GPAW, atomate, AiiDA & Battery cathodes/anodes, solid electrolytes, perovskite PV absorbers, electrocatalysts, thermoelectrics, permanent magnets, hydrides, corrosion alloys, fusion alloys & Materials Project, AFLOW, and OQMD demonstrate high-throughput first-principles screening \citep{jain2013materialsproject,curtarolo2012aflow,kirklin2015oqmd}; atomate automates first-principles workflows \citep{mathew2017atomate}. \scale{} can calibrate electronic-structure outputs against measured performance and distill them into composition/structure transformers. & Hours--days per structure; $10^4$--$10^7$ core-hours per campaign. \\
Finite-temperature atomistic transport & AIMD, NEB, CP2K, VASP-AIMD, Quantum ESPRESSO CP, VTST & Solid electrolytes, hydride diffusion pathways, perovskite ion migration, catalytic surfaces, radiation-defect migration & First-principles dynamics and migration-barrier workflows provide mechanistic transport labels. \scale{} can combine these labels with measured conductivity, diffusion, or degradation data for rapid transport screening. & Hours--days per path; $10^3$--$10^6$ core-hours per campaign. \\
Classical and reactive molecular dynamics & LAMMPS, GROMACS, OpenMM, AMBER, CHARMM, NAMD, ReaxFF & Polymer electrolytes, PEM/AEM membranes, MOFs, CO$_2$ sorbents, SEI/CEI layers, catalyst-ionomer films, phase-change materials, thermal-interface materials & Molecular simulation has enabled large adsorption and transport studies in porous materials \citep{wilmer2012mofs}. \scale{} can convert hydration, uptake, transport, swelling, morphology, and degradation simulations into calibrated transformer labels. & Hours--days per system; $10^3$--$10^6$ core-hours per campaign. \\
Machine-learned interatomic potentials & M3GNet, CHGNet, MatterSim-style potentials, DeePMD-kit, MACE, NequIP, GAP & Batteries, hydrides, high-entropy alloys, catalysts, perovskites, MOFs, fusion alloys, thermoelectrics & M3GNet and CHGNet show broad learned-potential transfer \citep{chen2022m3gnet,deng2023chgnet}; large atomistic models further broaden this direction \citep{yang2024mattersim}. \scale{} can place experimental calibration above ML-potential simulations. & High training cost; low--moderate inference cost per trajectory. \\
Thermodynamics and phase equilibria & Thermo-Calc, Pandat, FactSage, OpenCalphad, pycalphad & Hydrides, high-entropy alloys, molten salts, battery alloys, fusion blankets, nuclear alloys, thermoelectrics, phase-change alloys & pycalphad provides open CALPHAD infrastructure \citep{otis2017pycalphad}. \scale{} can learn calibrated phase-stability, phase-fraction, and plateau-pressure response surfaces. & Seconds--minutes per query after database assessment. \\
Reaction kinetics and kinetic Monte Carlo & Cantera, CHEMKIN, COPASI, Zacros, custom KMC solvers & Polymerization, electrolyte aging, CO$_2$ reduction, OER/HER/ORR catalysts, ammonia cracking, liquid organic hydrogen carriers & Zacros supports lattice KMC for catalytic surfaces \citep{nielsen2013zacros}. \scale{} can convert mechanism-based simulations and reaction experiments into fast reaction-outcome models. & Seconds--hours per ODE case; $10^2$--$10^5$ core-hours for KMC. \\
Mesoscale microstructure evolution & MOOSE, FiPy, FEniCS, PRISMS-PF, MICRESS & Battery cracking, lithium plating/dendrites, solid-state battery contact loss, hydride phase transformations, PCM solidification, perovskite grain growth, alloy precipitation, radiation damage & Phase-field and mesoscale solvers model microstructure-property evolution. \scale{} can learn calibrated morphology-to-property maps from simulation, microscopy, and operando data. & Hours--days per 2D case; $10^2$--$10^6$ core-hours for 3D. \\
Continuum mechanics and multiphysics & COMSOL, Abaqus, ANSYS, FEniCS, deal.II, MOOSE & Membrane-electrode contact, membrane swelling, solid-state battery stacks, PV module stress, fusion first-wall/divertor structures, power electronics packages & Contact-pressure studies are central to electrochemical stacks \citep{lin2009pemfcpressure}. \scale{} can amortize nonlinear stress/contact and coupled thermo-mechanical sweeps. & Minutes--hours simple; days for nonlinear 3D cases. \\
Electrochemical device simulation & PyBaMM, Dualfoil, MPET, PETLION, COMSOL electrochemistry & Li/Na batteries, flow batteries, fuel cells, electrolyzers, porous catalyst layers, gas diffusion layers & Doyle--Fuller--Newman theory underpins lithium-ion device simulation \citep{doyle1993dfn}; PyBaMM enables open battery modeling \citep{sulzer2021pybamm}. \scale{} can learn calibrated cell-level maps over materials, design, and operating windows. & ms--minutes reduced; hours--days for coupled/degradation cases. \\
Porous transport, CFD, and thermal systems & OpenFOAM, ANSYS Fluent, STAR-CCM+, COMSOL CFD, OpenPNM & Metal-hydride beds, carbon-capture packed beds, electrolyzer porous transport layers, fuel-cell gas diffusion layers, thermal runaway barriers, PCM thermal storage & OpenPNM supports pore-network transport modeling \citep{gostick2016openpnm}. \scale{} can learn calibrated transport/reactor fields and design windows. & Hours simple; days--weeks for 3D multiphase/reactive cases. \\
\bottomrule
\end{tabularx}
\end{table*}

\section{Implementation and Results}

We apply \scale{} to solid-state metal-hydride hydrogen-storage capacity prediction, a thermodynamic simulation and calibration problem aligned with the electronic-structure, phase-equilibrium, and finite-temperature modeling regimes in Table~\ref{tab:scale_categories}. Hydrogen is a strategically important energy carrier because it can connect low-cost renewable generation with industrial demand centers, long-duration storage, and chemical production, but practical deployment depends strongly on storage density, cost, safety, reversibility, kinetics, and operating temperature \citep{iea2025hydrogen,irena2022hydrogen,abdin2020hydrogen}. Solid-state metal hydrides offer high volumetric density and lower-pressure storage, yet their discovery remains constrained by coupled requirements on gravimetric H$_2$ capacity, plateau pressure, thermal management, cycling stability, elemental abundance, and manufacturability \citep{schlapbach2001hydrogen,zuttel2003materials,hirscher2020materials}. This implementation therefore tests whether \scale{} can convert a costly, mechanism-guided hydride-capacity workflow and limited measured anchors into calibrated training evidence for rapid materials inference. The following sections present the hydride capacity operator, experimental calibration, calibrated dataset construction, graph preparation, transformer learning, and surrogate-fidelity evaluation.

\subsection{Hydride Phase-Equilibrium Capacity Operator}

The implementation starts from version 0.0.3 of the open ML-HydPARK dataset, formally titled \emph{Database for machine learning of hydrogen storage materials properties}, derived from the U.S. DOE HydPARK resource and released by Witman, Allendorf, and Stavila \citep{witman2022mlhydpark}. We use Zenodo record DOI 10.5281/zenodo.7324807, file \texttt{ML-HYDPARK\_v0.0.3.csv}. The downloaded table contains 429 rows and 12 columns, including material class, alloy composition, measured H$_2$ wt\%, absorption temperature, hydrogen pressure, reaction enthalpy and entropy, equilibrium pressure at 25 $^\circ$C, H/M ratio, and source reference. For the present benchmark, only rows with alloy composition, absorption temperature, hydrogen pressure, and measured H$_2$ wt\% are eligible; after formula and operator-input validation, 381 complete rows are used for calibration. Later ML-HydPARK releases exist, including Zenodo record 10.5281/zenodo.10680097, version 0.0.5, but they are not mixed into the results reported here. The dataset spans intermetallic and alloy families including A$_2$B, AB, AB$_2$, AB$_5$, Mg-based alloys, miscellaneous intermetallic compounds, and solid-solution systems.

The hydride phase-equilibrium capacity operator is constructed as follows.

Each candidate is represented by composition, hydride class, and operating condition:
\begin{align}
  r_i = (f_i,\kappa_i,T_i,P_{i,\mathrm{H_2}}),
  \label{eq:mh_input}
\end{align}
where $f_i$ is the hydrogen-free parent-alloy formula, $\kappa_i$ is the hydride class, $T_i$ is the absorption temperature, and $P_{i,\mathrm{H_2}}$ is the hydrogen pressure. The deterministic model uses this deployable candidate record and reconstructable structural information.

The hydride phase-equilibrium capacity operator converts $r_i$ into a finite-temperature hydrogen-loading problem. It parses the parent-alloy formula, computes elemental fractions and molar mass, assigns or verifies the hydride class, resolves a parent-alloy structure from Materials Project, OQMD, or class-consistent prototypes, relaxes the parent structure, analyzes candidate interstitial environments, enumerates hydrogen-loading hypotheses, screens hydrogenated branches, promotes selected branches to full first-principles confirmation, compares the condition-dependent free-energy ladder, and selects the accessible hydrogen loading by mass balance \citep{ong2013pymatgen,jain2013materialsproject,kirklin2015oqmd,giannozzi2009quantum,togo2024spglib}.

For a parent alloy $M$ and a hydrogenated candidate phase $MH_x$, the operator evaluates condition-dependent hydride stability:
\begin{align}
  \Delta G_x(T,P_{\mathrm{H_2}})
  =
  G(MH_x,T)-G(M,T)-\frac{x}{2}\mu_{\mathrm{H_2}}(T,P_{\mathrm{H_2}}).
  \label{eq:mh_delta_g}
\end{align}
The stable or accessible loading is selected by minimizing the free-energy ladder:
\begin{align}
  x_i^\star=\arg\min_x \Delta G_x(T_i,P_{i,\mathrm{H_2}}),
  \label{eq:mh_xstar}
\end{align}
subject to phase-stability, structural-quality, and reversibility checks. The raw deterministic gravimetric capacity is then calculated by mass balance:
\begin{align}
  y_{i,\mathrm{op}}
  =100\times\frac{x_i^\star M_{\mathrm{H}}}{M_{\mathrm{alloy},i}+x_i^\star M_{\mathrm{H}}},
  \label{eq:mh_wt}
\end{align}
where $M_{\mathrm{H}}$ is the atomic mass of hydrogen and $M_{\mathrm{alloy},i}$ is the molar mass of the parent alloy formula.

The deterministic operator returns a structured state vector:
\begin{align}
  \bm{s}_i =
  [y_{i,\mathrm{op}},x_i^\star,\Delta G_i(x_1),\ldots,\Delta G_i(x_B),\Delta V_i,\eta_{\mathrm{stab},i},\mathrm{QC}_i],
  \label{eq:mh_state}
\end{align}
where $y_{i,\mathrm{op}}$ is the raw H$_2$ wt\% estimate, $x_i^\star$ is the selected hydrogen loading, the $\Delta G$ terms define the free-energy ladder, $\Delta V$ records volume or structure change on hydrogenation, $\eta_{\mathrm{stab}}$ is a phase-stability margin, and $\mathrm{QC}$ stores convergence and structural-validity flags. Table~\ref{tab:mh_det_workflow} gives the deterministic workflow used to construct the operator output supplied to calibration.

\begin{table*}[t]
\centering
\caption{Hydride phase-equilibrium capacity operator used in the metal-hydride implementation. The workflow begins with the hydrogen-free parent alloy, enumerates hydrogen-loading hypotheses without using measured capacity or known hydride stoichiometry, and promotes selected branches to full first-principles confirmation. Computational-burden expressions describe deterministic work rather than hardware-specific wall time. Because the present implementation evaluates the SCALE architecture rather than establishing a definitive metal-hydride phase diagram, Step 7 was benchmarked on representative candidate branches to quantify the computational burden of terminal variable-cell confirmation but was not executed exhaustively across the expanded corpus. Note: $N$ is the number of atoms in the quantum cell, $K$ is the number of irreducible $k$-points, $I_{\mathrm{scf}}$ is the number of electronic iterations per geometry step, $I_{\mathrm{geo}}$ is the number of ionic or cell-relaxation steps, $B$ is the hydrogen-loading ladder size, and $m$ is the number of branches promoted to full relaxation. Heavy elements, semicore states, metallic smearing, low symmetry, and large ordered approximants can substantially increase the computational prefactor.}
\label{tab:mh_det_workflow}
\scriptsize
\renewcommand{\arraystretch}{1.08}
\setlength{\tabcolsep}{1.1pt}
\begin{tabularx}{\textwidth}{L{0.035\textwidth}L{0.105\textwidth}L{0.135\textwidth}L{0.145\textwidth}L{0.255\textwidth}L{0.255\textwidth}}
\toprule
Step & Stage & Input & Output & Tools and method & Computational burden \\
\midrule
1 & Parent-alloy definition & Formula, material class, $T$, and $P_{\mathrm{H_2}}$ & Parsed composition, molar mass, and condition record & \texttt{pymatgen} and validated composition parsing \citep{ong2013pymatgen} & Low; linear in the number of elements, $O(E)$. \\
2 & Parent-structure resolution & Parsed formula and parent-only structure sources & Exact parent CIF or ordered parent approximant & Materials Project or OQMD retrieval; deterministic supercell and sublattice substitution \citep{jain2013materialsproject,kirklin2015oqmd,ong2013pymatgen} & Low--moderate; API/cache dominated for exact structures, increasing with approximant size. \\
3 & Parent quantum relaxation & Parent CIF, pseudopotentials, $k$-mesh, cutoffs, and smearing & Relaxed or plateau-accepted parent geometry and energy & Quantum ESPRESSO variable-cell relaxation with PBE pseudopotentials \citep{giannozzi2009quantum} & High; approximately $C_{\mathrm{vc}}(N,K,I)\sim N^{2.5\text{--}3}K I_{\mathrm{scf}}I_{\mathrm{geo}}$. \\
4 & Parent-structure analysis & Relaxed parent geometry & Distances, symmetry/prototype, and candidate interstitial regions & \texttt{pymatgen}, \texttt{spglib}, periodic distance analysis, and Voronoi/grid void search \citep{ong2013pymatgen,togo2024spglib} & Low--moderate; pair analysis $O(N^2)$, with void-search cost set by grid density. \\
5 & Hydrogen-loading hypotheses & Candidate interstices, formula units, and collision constraints & Parent+H structures across discrete loading states & Deterministic H-site placement, loading enumeration, and structure/input writers & Moderate; scales with candidate sites and loading branches, approximately $O(BN)$. \\
6 & Fast quantum screening & Parent+H candidates and parent/H$_2$ references & Ranked loading candidates & Fixed-cell or short-relaxation Quantum ESPRESSO screening, energy parsing, and formation-energy ranking & High, but screening grade; approximately $B\,C_{\mathrm{scf}}(N_H,K,I)$. \\
7 & Full hydride confirmation and capacity selection & Top $m$ candidates from Step 6, normally $m=3$ & Final H$_2$ wt\%, relaxed hydrides, energy, and QC record & Full Quantum ESPRESSO variable-cell relaxation, free-energy comparison, phase selection, and mass balance & Very high; approximately $m\,C_{\mathrm{vc}}(N_H,K,I)$ after parent relaxation and branch screening. \\
\bottomrule
\end{tabularx}
\end{table*}
\renewcommand{\arraystretch}{1.0}

\subsection{Experimental Calibration of the Hydride Capacity Operator}

The hydride phase-equilibrium capacity operator supplies a structured state $\bm{s}_i$ for each candidate-condition record $r_i$, including the predicted capacity, selected hydrogen loading, free-energy ladder, structural response, phase-stability margin, and quality-control indicators. These mechanistic outputs need calibration because measured capacity is also influenced by synthesis route, phase purity, crystallinity, activation history, hysteresis, heat and mass transfer, and incomplete reversibility.

We fit the calibration model using 381 validated ML-HydPARK records. Inputs are restricted to deployable candidate descriptors and quantities generated by the deterministic operator; measured H$_2$ wt\% serves only as the calibration target. Database fields containing near-target thermodynamic information are excluded unless independently generated by the operator, thereby preventing target leakage.

Following the model-selection criterion defined in Section 2.1, we evaluated regularized linear regression, histogram-gradient boosting, random forests, and extremely randomized trees \citep{geurts2006extratrees,pedregosa2011sklearn}. Extremely randomized trees were selected because they achieved the lowest random five-fold MAE. For this implementation, the calibrated response is written as
\begin{align}
  y_i^{\mathrm{cal}}
  =
  y_{i,\mathrm{op}}
  +
  \delta_{\phi^\star}(r_i,\bm{s}_i),
  \label{eq:mh_residual}
\end{align}
where $\delta_{\phi^\star}$ is the learned residual correction implied by the selected calibrator.

Table~\ref{tab:mh_calibration_results} summarizes calibration performance. The uncalibrated operator retains a substantial monotonic relationship with experiment ($r=0.823$) but underestimates capacity by an average of $0.584$ wt\% H$_2$. Calibration reduces the raw-operator MAE of $0.623$ wt\% H$_2$ to a random five-fold MAE of $0.219$ wt\% H$_2$. The grouped splits provide more stringent tests of transfer across material families, temperature regimes, and dominant-element chemistries.

\begin{table*}[t]
\centering
\caption{Experimental calibration accuracy against ML-HydPARK measured H$_2$ wt\%. The selected calibration uses deployable chemistry/condition descriptors and deterministic-operator outputs; raw operator performance directly compares uncalibrated simulation-derived capacity to experiment.}
\label{tab:mh_calibration_results}
\small
\setlength{\tabcolsep}{3.1pt}
\begin{tabular}{llccccc}
\toprule
Model & Validation split & $n$ & MAE & RMSE & $R^2$ & Pearson $r$ \\
\midrule
Hydride capacity operator & all rows & 381 & 0.623 & 0.788 & 0.274 & 0.823 \\
Ridge calibration & random 5-fold & 381 & 0.291 & 0.450 & 0.763 & 0.875 \\
Histogram-gradient boosting & random 5-fold & 381 & 0.230 & 0.371 & 0.839 & 0.916 \\
Random forest & random 5-fold & 381 & 0.234 & 0.367 & 0.843 & 0.918 \\
Extremely randomized trees & random 5-fold & 381 & \textbf{0.219} & 0.380 & 0.831 & 0.912 \\
Extremely randomized trees & material-class grouped & 381 & 0.405 & 0.610 & 0.565 & 0.761 \\
Extremely randomized trees & temperature-bin grouped & 381 & 0.318 & 0.485 & 0.725 & 0.858 \\
Extremely randomized trees & dominant-element grouped & 381 & 0.511 & 0.769 & 0.309 & 0.634 \\
\bottomrule
\end{tabular}
\end{table*}

The resulting labels therefore preserve the mechanistic structure of the deterministic operator while correcting its systematic deviation from measured capacity. These fidelity-calibrated labels provide the teaching evidence used for expanded dataset construction.

\subsection{Calibrated Virtual-Label Generation}

After calibration, the calibrated operator is used offline to populate a larger training set. The 5,000-row virtual candidate set is generated from deployable design variables: alloy composition, hydride class, prototype scaffold, absorption temperature, and hydrogen pressure. Candidate compositions are generated by class-aware perturbation of ML-HydPARK anchor compositions, while temperature and pressure are sampled within class-supported operating ranges. Each row is assigned a prototype scaffold corresponding to its class: A$_2$B intermetallic, AB ordered intermetallic, AB$_2$ Laves-like, AB$_5$ CaCu$_5$-like, Mg-rich light-metal hydride, miscellaneous intermetallic compound (MIC), or substitutional solid-solution (SS) scaffold. The virtual table contains 314 A$_2$B, 581 AB, 1103 AB$_2$, 998 AB$_5$, 617 Mg, 711 MIC, and 676 SS candidates. The raw deterministic H$_2$ wt\% range is 0.039--5.996, and the calibrated teacher range is 0.272--5.635.

For each virtual candidate $r_i$, the deterministic operator produces $\bm{s}_i$ and the calibration model assigns the teacher label
\begin{align}
  y_i^{F}=C_{\phi^\star}(r_i,\bm{s}_i),
  \qquad
  \mathcal{D}_{F}=\{(r_i,y_i^{F})\}_{i=1}^{5000}.
  \label{eq:mh_virtual_dataset}
\end{align}
The deployed model does not receive $\bm{s}_i$ or $y_{i,\mathrm{op}}$. Those quantities create the offline teacher labels, preserving the central amortization step: new candidate-condition queries do not rerun the deterministic workflow.

\subsection{Periodic Graph Generation for Metal-Hydride Candidates}

Formula- and composition-based encodings are common in materials machine learning because they are inexpensive, easy to standardize, and compatible with large chemical spaces. For crystalline metal hydrides, however, they compress the input into averaged elemental information and do not explicitly retain crystallographic site multiplicity, interatomic distances, periodic images, local coordination, or chemically distinct metal--metal environments. These structural quantities are relevant because hydrogen accommodation is governed by the geometry and chemistry of the host lattice, not only by the elements present \citep{xie2018cgcnn,chen2019megnet,choudhary2021alignn,yan2022matformer}.

For this implementation, we therefore use a parent-alloy periodic graph. Each candidate record is
\begin{align}
  r_i=(f_i,\kappa_i,T_i,P_{i,\mathrm{H_2}}),
  \label{eq:mh_graph_input}
\end{align}
where $f_i$ is the hydrogen-free parent-alloy formula, $\kappa_i$ is the material class, $T_i$ is the operating temperature, and $P_{i,\mathrm{H_2}}$ is the hydrogen pressure. The graph-preparation layer parses the composition, resolves a parent crystal structure from cached Materials Project records when available, and otherwise selects a hydrogen-free class-consistent prototype: LaNi$_5$ for AB$_5$, ZrMn$_2$ for AB$_2$, TiFe for AB, Mg$_2$Ni for A$_2$B and Mg-rich alloys, and TiCrMn for miscellaneous intermetallic or solid-solution-like entries. Candidate elements are substituted onto prototype sublattices using composition fractions and fixed chemistry-role conventions. This preserves the source unit-cell topology, fractional coordinates, lattice geometry, and periodic neighbor relations while adapting elemental identity to the candidate composition \citep{jain2013materialsproject,ong2013pymatgen}.

The resulting input is
\begin{align}
  G_i=(V_i,E_i,\bm{a}_i),
  \label{eq:mh_graph}
\end{align}
where $V_i$ is the set of parent-alloy atomic sites, $E_i$ is the set of directed periodic-neighbor edges, and $\bm{a}_i$ is a global descriptor vector containing operating conditions, composition statistics, lattice metrics, symmetry information, structure-source confidence, and material-class indicators. Each node is represented by an 18-dimensional feature vector containing elemental descriptors, formula fraction, fractional coordinates, parent-site role, and local coordination. Each edge is represented by a 22-dimensional feature vector containing periodic neighbor distance, radial-basis distance channels, elemental and hydride-affinity contrasts, metal--metal relation type, periodic image vector, lattice-length context, and electronegativity contrast.

\begin{figure*}[t]
\centering
\includegraphics[width=0.94\textwidth]{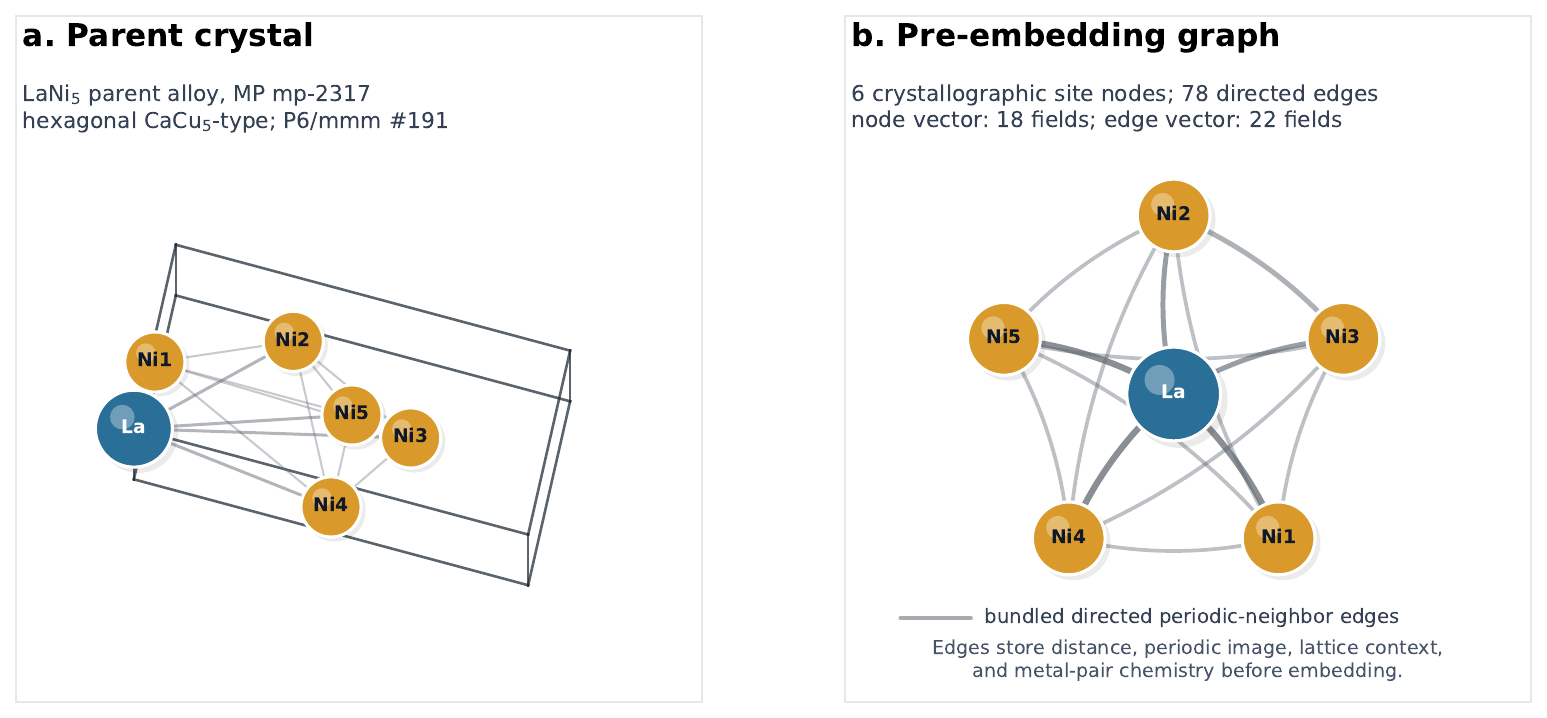}
\caption{Parent-alloy structure and periodic graph before transformer embedding. \textbf{a,} The crystallographic LaNi$_5$ parent cell, with metal--metal neighbor lines shown through the periodic structure. \textbf{b,} The corresponding model input graph, in which crystallographic sites become nodes and periodic neighbors become directed edges carrying geometric and chemistry descriptors. The graph is constructed from deployable parent-alloy information and contains no target-derived loading assignment.}
\label{fig:mh_structure_graph}
\end{figure*}

This preparation is a low-cost structural encoding step rather than a deterministic capacity calculation. It produced 5,000 valid parent-alloy graphs: 1,478 used cached parent-alloy Materials Project structures and 3,522 used class-prototype substitutions. The padded tensor dimensions are
\begin{align}
  \bm{X}_{V}&\in\mathbb{R}^{5000\times34\times18},
  \nonumber\\
  \bm{X}_{E}&\in\mathbb{R}^{5000\times352\times22},
  \nonumber\\
  \bm{X}_{A}&\in\mathbb{R}^{5000\times33},
  \label{eq:mh_prepared_tensor_dims}
\end{align}
for node features, directed-edge features, and global features, respectively.

\subsection{Graph Transformer on the Prepared Periodic Graph}

The prepared tensors in Eq.~\eqref{eq:mh_prepared_tensor_dims} are supplied with masks for variable site and edge counts. Node features are embedded from 18 to hidden width 192 by a two-layer projection block with layer normalization and SiLU activation, while the 33 global features are independently embedded to 192 dimensions. Six edge-biased graph-transformer layers then update the site representations. Each layer uses eight attention heads, head dimension 24, feed-forward width 768, dropout 0.08, residual connections, and pre-layer normalization. For head $h$, the attention update is
\begin{align}
  \bm{Q}^{h}=\bm{H}\bm{W}_{Q}^{h},\quad
  \bm{K}^{h}=\bm{H}\bm{W}_{K}^{h},\quad
  \bm{V}^{h}=\bm{H}\bm{W}_{V}^{h},
  \label{eq:mh_qkv}
\end{align}
\begin{align}
  \mathrm{Attn}^{h}_{ij}
  =
  \mathrm{softmax}_{j}
  \left(
  \frac{\bm{Q}^{h}_{i}\bm{K}^{hT}_{j}}{\sqrt{24}}
  +b^{h}_{ij}
  \right),
  \nonumber\\
  b^{h}_{ij}
  =
  \mathrm{MLP}_{h}(\bm{e}_{ij}),
  \label{eq:mh_edge_attention}
\end{align}
where $\bm{e}_{ij}$ is the 22-dimensional periodic-edge feature vector. The learned edge bias conditions information aggregation on neighbor distance, periodic image, metal-role relation, and chemistry contrasts rather than treating all crystallographic neighbors as equivalent. The final transformer layer is summarized by a learned gated attention pool:
\begin{align}
  \alpha_{ij}
  &=
  \mathrm{softmax}_{j}
  [\bm{w}_{p}^{T}\sigma(\bm{W}_{p}\bm{h}_{ij})],
  \nonumber\\
  \bm{g}_{i}
  &=
  \sum_j \alpha_{ij}\bm{h}_{ij}.
  \label{eq:mh_pool}
\end{align}
The graph embedding $\bm{g}_i$ is concatenated with the 192-dimensional global embedding, and a $384\rightarrow192\rightarrow96\rightarrow1$ regression head with GELU activations produces the capacity estimate:
\begin{align}
  \hat{y}_{i}^{F}
  =
  R_{\theta}
  ([\bm{g}_i,\mathrm{MLP}_{a}(\bm{a}_i)]).
  \label{eq:mh_regression}
\end{align}
The trainable model contains 2,901,170 parameters. Training uses Smooth L1 loss with transition parameter 0.35 on normalized calibrated H$_2$ wt\% labels, AdamW optimization, initial learning rate $4\times10^{-4}$, weight decay $2\times10^{-4}$, cosine learning-rate annealing, batch size 96, validation fraction 0.12 inside each fold, and 60 training epochs. Random five-fold evaluation, initialized from seed 20260829, measures surrogate fidelity over the populated teacher distribution.

\subsection{Calibration and Transformer Results}

\begin{figure*}[t]
\centering
\includegraphics[width=0.94\textwidth]{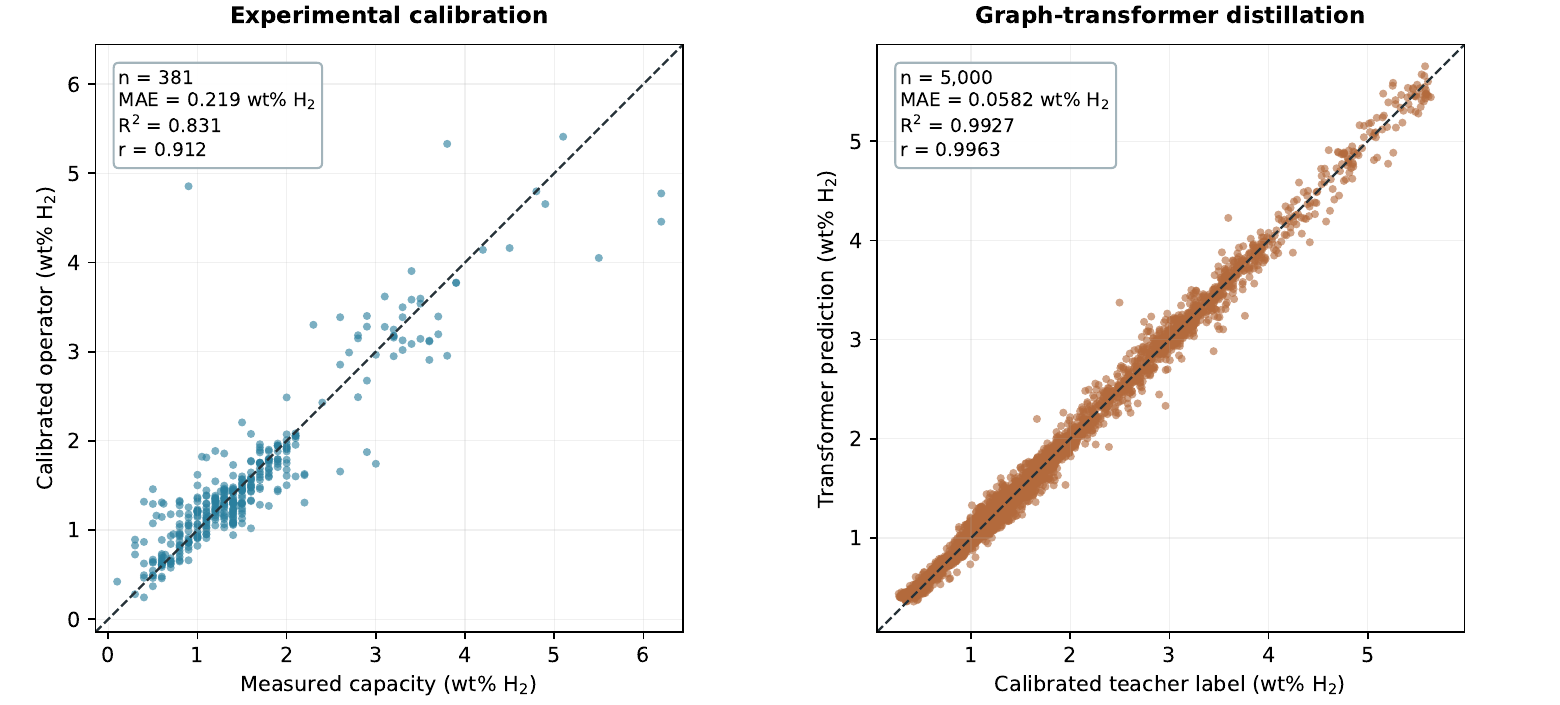}
\caption{Parity plots for the two supervised links in the metal-hydride implementation. Left: experimental calibration model predictions against measured H$_2$ wt\% values. Right: graph-transformer predictions against held-out calibrated teacher labels in the 5,000-row virtual dataset.}
\label{fig:mh_parity}
\end{figure*}

Figure~\ref{fig:mh_parity} shows the two supervised mappings in the implementation: experimental calibration and transformer distillation. The calibration panel measures agreement with ML-HydPARK experimental anchors under random five-fold validation. The transformer panel measures held-out agreement with calibrated teacher labels in the 5,000-row virtual dataset.

\begin{table*}[t]
\centering
\caption{Overall and fold-level graph-transformer surrogate fidelity against calibrated metal-hydride H$_2$ wt\% teacher labels. The overall row reports metrics calculated from all 5,000 out-of-fold predictions; fold rows demonstrate consistency across independently trained models.}
\label{tab:mh_transformer_metrics}
\small
\setlength{\tabcolsep}{4.0pt}
\begin{tabular}{lcccccc}
\toprule
Split & $n$ & MAE & RMSE & $R^2$ & Pearson $r$ & Bias \\
\midrule
Overall OOF & 5000 & 0.0582 & 0.0833 & 0.9927 & 0.9963 & $+0.0022$ \\
Fold 1 & 1000 & 0.0593 & 0.0901 & 0.9919 & 0.9960 & $+0.0080$ \\
Fold 2 & 1000 & 0.0591 & 0.0849 & 0.9929 & 0.9964 & $-0.0017$ \\
Fold 3 & 1000 & 0.0582 & 0.0825 & 0.9920 & 0.9960 & $+0.0021$ \\
Fold 4 & 1000 & 0.0589 & 0.0810 & 0.9932 & 0.9966 & $-0.0026$ \\
Fold 5 & 1000 & 0.0556 & 0.0773 & 0.9934 & 0.9967 & $+0.0054$ \\
\bottomrule
\end{tabular}
\end{table*}

These metrics measure whether the graph transformer can reproduce the calibrated teacher response surface from deployable hydride inputs. They do not claim that the transformer independently supersedes experimental validation. The purpose of this stage is amortization: once the hydride capacity operator and experimental calibration model have populated a teacher distribution, the transformer can answer new candidate-condition queries without rerunning the deterministic phase-equilibrium workflow.

\par\medskip
\subsection{Compute and Reproducibility}

The implementation package records dataset audit notes, deterministic-workflow documentation, calibration artifacts, virtual-dataset construction scripts, graph-preprocessing scripts, transformer training scripts, trained checkpoints, cross-validation metrics, inference benchmarks, and provenance notes. The transformer benchmark reports five independently trained rotating folds rather than a selected split; every out-of-fold prediction is generated by a model that did not train on that row. All folds completed 60 epochs. In a cached CPU benchmark using batch size 256, one checkpoint processed 5,000 prepared graphs in 9.24 s, corresponding to approximately 541 candidates s$^{-1}$.

\section{Discussion}

\subsection{Transformer Learning from Calibrated Operator Results}

The metal-hydride implementation demonstrates the central learning mechanism of SCALE. The 381 ML-HydPARK anchors are sufficient to calibrate the structured outputs of the hydride capacity operator, but not to cover the alloy, prototype, temperature, and pressure design space by experiment alone. Repeating the deterministic operator for every new candidate would require parent-structure resolution, parent relaxation, interstitial analysis, hydrogen-loading enumeration, branch screening, selected hydride confirmation, free-energy comparison, and mass-balance capacity readout. SCALE instead uses this calibrated operator workflow to construct a finite teacher distribution, then trains the graph transformer to predict calibrated H$_2$ capacity from deployable parent-alloy graph, composition/class, temperature, and pressure inputs.

The transformer results are therefore best read as a surrogate-fidelity test: the model is not being evaluated as a direct substitute for experiment, but as an amortized learner of the calibrated operator response surface. The low held-out error and high agreement reported in Section~3.6, with MAE $0.0582$ wt\% H$_2$ and $R^2=0.9927$ over 5,000 out-of-fold teacher-labeled candidates, indicate that the prepared periodic graph representation and edge-biased transformer architecture preserve the teacher distribution with sufficient accuracy for ranking, triage, and large-scale candidate screening.

The remaining scientific uncertainty resides in auditable upstream layers: operator construction, experimental-anchor quality, calibration transfer, and virtual-candidate coverage. These layers define the validation strategy discussed in Section~4.4.

\subsection{Attention-Derived Chemical Organization}

Whether the trained SCALE graph transformer develops chemically organized internal representations was evaluated through post hoc analysis of learned attention structure. Element-level gate enrichment and directed element-pair attention enrichment were extracted from the five out-of-fold models, followed by unsupervised clustering and ranking of the learned attention fingerprints. The analysis used only model-derived attention statistics; chemical interpretation was applied afterward.

The attention-derived organization showed reproducible structure. Element-level clustering repeatedly produced a high-enrichment group containing Ni, Co, Fe, Mn, Pt, Ir, and Ag, while lower-enrichment or structure-bearing groups more often contained La, Ce, Nd, Y, Mg, Ti, and Zr. Quantitatively, the high-enrichment group had greater average gate enrichment than the remaining elements; a simple attention-enrichment threshold recovered this behavior with precision 0.56, recall 0.82, and F1 score 0.67. Directed element-pair analysis identified 372 recurring element-pair classes with at least 50 observations. The highest-enrichment relationships included Er$\rightarrow$Co, Sm$\rightarrow$Co, Ho$\rightarrow$Fe, Ni$\rightarrow$Nd, Mn$\rightarrow$Mg/Ca, and Pd$\rightarrow$Co/Cr. Full blind clustering of the element-pair matrix reached a maximum purity of approximately 0.48, indicating partial rather than complete separation.

This unsupervised attention organization is highly consistent with established chemical grouping and metal-hydride science. The high-enrichment set is dominated by late or related transition metals, including Ni, Co, Fe, Mn, Pt, Ir, and Ag, whereas the lower-enrichment or structure-bearing set contains many rare-earth, light-metal, and early-transition-metal hydride hosts, including La, Ce, Nd, Y, Mg, Ti, and Zr. Hydrogen capacity and hydride stability are governed not only by average formula, but also by host-lattice geometry, interstitial-site availability, local metal-metal environments, and alloying elements that tune hydrogen-binding thermodynamics. Rare-earth and light-metal hosts provide hydrogen-accommodating frameworks in many hydride families, while transition metals such as Ni, Co, Fe, Mn, Pd, and related elements strongly influence phase stability, electronic structure, and absorption/desorption behavior \citep{schlapbach2001hydrogen,pasquini2022magnesium,dematteis2021substitutional}. The emergence of these relationships in attention-derived groupings, without being imposed as the clustering target, supports the view that SCALE learns chemically structured representations rather than only scalar correlations.

These findings provide evidence for the broader SCALE premise: calibrated scientific evidence can be transferred into advanced learning architectures while retaining interpretable links to domain principles. In this implementation, the graph-transformer design enables relational scientific structure to become part of the learned representation, strengthening confidence that fast SCALE inference is not only accurate but scientifically organized.

\begin{figure*}[t]
\centering
\includegraphics[width=\textwidth]{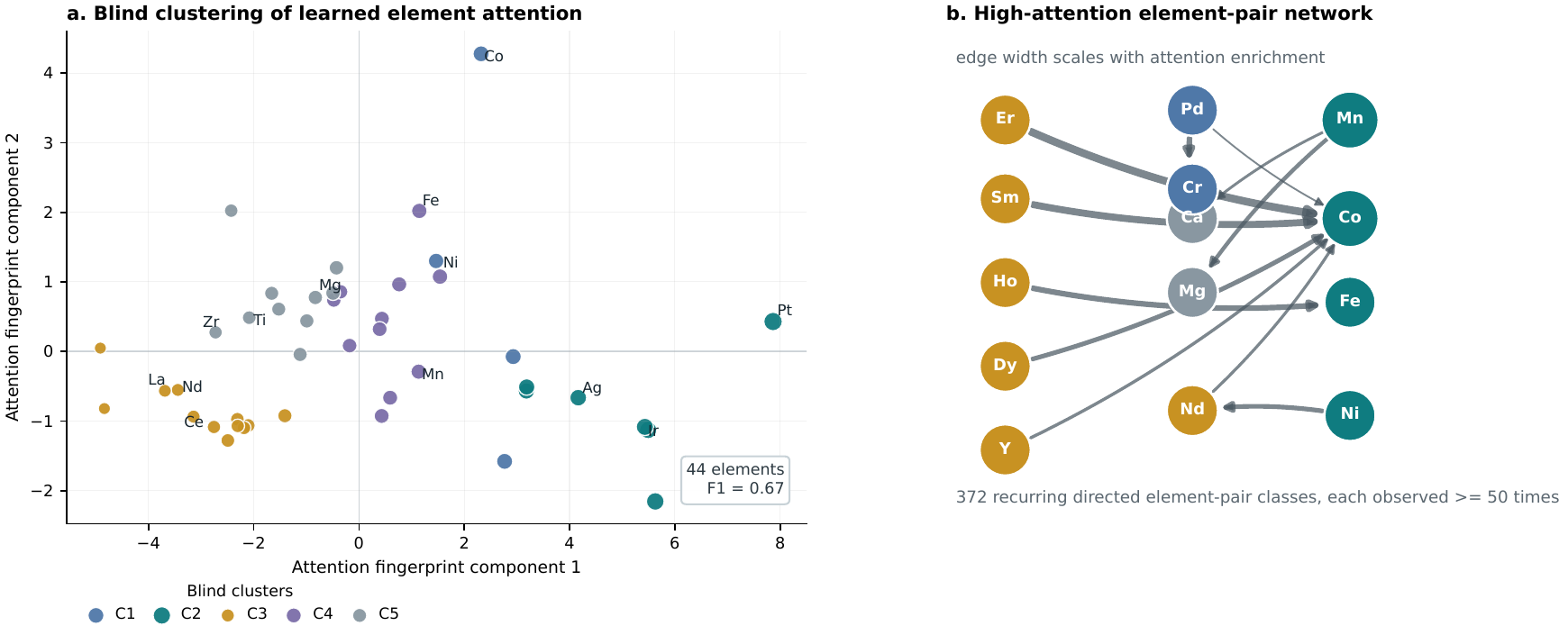}
\caption{Attention-derived chemical organization in the SCALE metal-hydride implementation. \textbf{a,} Unsupervised organization of element-level attention fingerprints extracted from the five out-of-fold graph-transformer models. Each point represents one element with at least 50 node observations; colors denote blind clusters obtained from attention statistics only. Selected element labels are shown for readability. The high-enrichment region contains Ni, Co, Fe, Mn, Pt, Ir, and Ag, while lower-enrichment or structure-bearing regions contain La, Ce, Nd, Y, Mg, Ti, and Zr. A post hoc attention-enrichment threshold recovers this separation with F1 = 0.67. \textbf{b,} Directed element-pair attention network for recurring high-enrichment relationships. Nodes are elements and arrows denote directed attention allocation between element sites; edge width scales with mean attention enrichment. The displayed relationships are selected from 372 directed element-pair classes observed at least 50 times. The network highlights rare-earth/transition-metal, host/transition-metal, and transition-metal/transition-metal relationships that are consistent with established metal-hydride chemistry.}
\label{fig:mh_attention}
\end{figure*}

\subsection{Speed and Financial Scaling}

For a campaign with $N$ candidates, repeated deterministic screening scales as
\begin{align}
  T_{\mathrm{det}}(N)=\frac{N t_{\mathrm{op}}}{P_{\mathrm{eff}}},
  \label{eq:mh_det_time}
\end{align}
where $t_{\mathrm{op}}$ is the wall time per candidate-condition deterministic workflow and $P_{\mathrm{eff}}$ is effective parallelism after queueing, utilization, failure recovery, and workflow overhead. Transformer inference scales as
\begin{align}
  T_{\mathrm{tr}}(N)=T_{\mathrm{prep}}(N)+\frac{N}{R_{\mathrm{infer}}},
  \label{eq:mh_transformer_time}
\end{align}
where $T_{\mathrm{prep}}$ is graph-feature preparation and $R_{\mathrm{infer}}$ is batched model throughput.

A practical hydride phase-equilibrium campaign can require many structure candidates and hydrogen-loading branches per alloy. Using 12--72 h per candidate as a planning range for prototype generation, relaxation, hydrogenated-structure enumeration, finite-temperature correction, and readout, one million candidates imply $1.2\times10^7$--$7.2\times10^7$ deterministic candidate-hours. At an illustrative scientific-computing cost of US\$0.75--US\$3.00 per worker-hour, the deterministic-only compute bill would be approximately US\$9--US\$216 million for the million-candidate screen, before failed jobs, queue delays, reruns, and expert review.

\begin{figure*}[t]
\centering
\includegraphics[width=\textwidth]{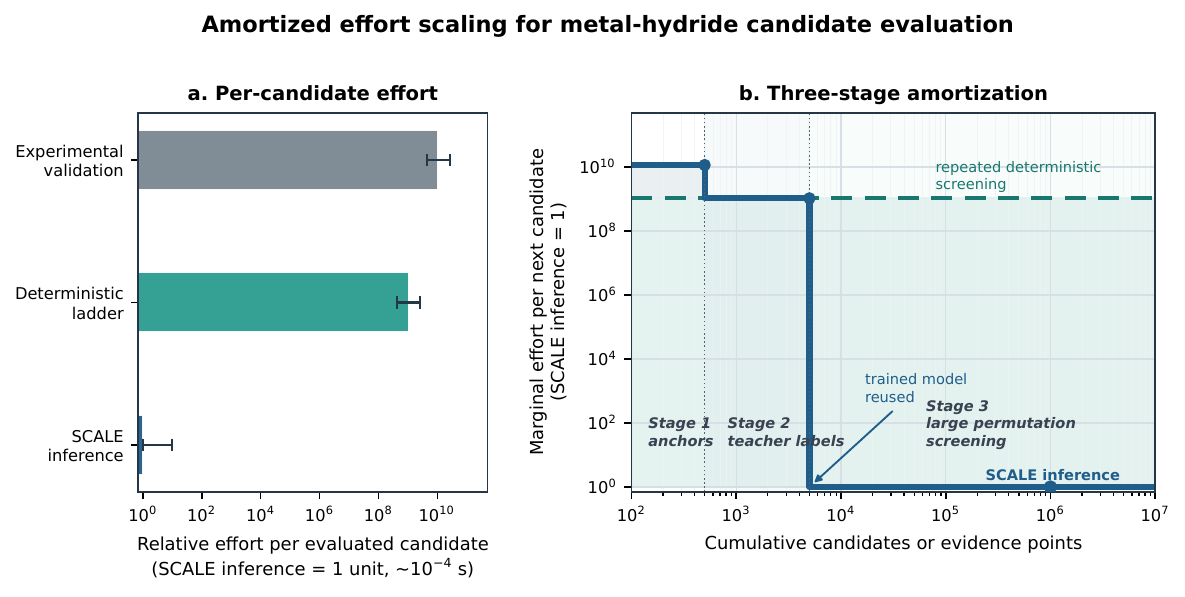}
\caption{Normalized effort and amortization scaling for metal-hydride candidate evaluation. \textbf{a,} Per-candidate effort normalized to one trained SCALE inference call. Experimental validation represents synthesis, activation, and capacity measurement for a candidate material; the deterministic ladder represents parent-alloy relaxation, hydrogen-loading enumeration, branch screening, and selected hydride refinement; SCALE inference represents one deployed graph-transformer forward pass after training. Error bars indicate planning ranges rather than hardware-independent constants. \textbf{b,} Three-stage amortization model. Stage 1 uses experimental anchors together with deterministic modeling and calibration to establish calibrated evidence. Stage 2 populates a larger teacher-labeled set through the calibrated deterministic workflow. Stage 3 reuses the trained SCALE model for large permutation screening, causing the marginal effort per additional candidate to collapse toward inference cost. The y-axis is normalized to SCALE inference $=1$ unit; as a practical reference, one unit corresponds to an assumed order of $10^{-4}$ s per candidate on a high-end data-center GPU, while deterministic-ladder estimates use 12--72 h per candidate workflow and experimental validation uses 5--30 days per sample. These values are planning-scale estimates; wall time depends on hardware, queueing, parallelism, convergence behavior, and experimental protocol.}
\label{fig:mh_cost}
\end{figure*}

The trained graph transformer changes this scaling. The cost of teacher generation is still paid for the selected training set, but it is no longer paid for every screened candidate. In the measured cached CPU benchmark, one checkpoint processed approximately 541 candidates s$^{-1}$, corresponding to about 30.8 min for one million candidates. For normalized planning analysis, Figure~\ref{fig:mh_cost} expresses effort relative to one trained SCALE inference call and uses an order-of-magnitude high-end accelerator reference of $10^{-4}$ s per candidate.

This difference is not merely a speed claim. It changes the design loop. Deterministic models and experiments are reserved for selected high-value points: calibration anchors, uncertainty reduction, frontier candidates, and mechanistic audits. The transformer carries the high-throughput burden of searching the combinatorial design space.

\subsection{General Validation Strategy for \scale{}}

Validation of \scale{} should be organized by evidence layer rather than by a single aggregate accuracy number. Deterministic-operator validation verifies that the scientific workflow solves the intended mechanistic problem with appropriate convergence, boundary conditions, structural enumeration, thermodynamic references, and selection criteria. Calibration validation tests whether experimental anchors correct systematic operator bias without leakage across train/test splits. Teacher-distribution validation evaluates whether expanded calibrated labels remain inside the scientific domain where the operator and experiments are credible. Transformer-fidelity validation measures whether the deployed model reproduces the calibrated teacher distribution with low numerical and ranking error. Finally, prospective validation closes the loop by sending selected high-value or high-uncertainty candidates back to deterministic modeling and experiment. Across \scale{} regimes, this layered validation separates surrogate accuracy from scientific discovery claims while identifying where new computation and measurement have the highest value.

\subsection{Limitations}

\scale{} does not remove the need for expert deterministic modeling or high-quality experiments. It reorganizes when they are paid for. The transformer inherits the domain boundaries of the calibrated teacher. If the hydride operator misses a mechanism and the experimental calibration set does not expose it, the transformer can reproduce that blind spot at scale. Similarly, if the virtual candidate generator leaves the realistic synthesis domain, the model can make precise-looking predictions for candidates that are not chemically or manufacturably credible.

The present metal-hydride implementation also shows why validation splits matter. Random-row calibration and transformer folds quantify interpolation over the populated teacher distribution. Material-class, temperature-bin, and dominant-element grouped splits are more stringent tests of transfer. Dominant-element grouped calibration remains the weakest test in the present results, indicating that extrapolation to chemically distinct elemental families is a principal residual risk. This observation is not a failure of amortization; it identifies where new deterministic calculations and experimental anchors would have the highest marginal value.

The metal-hydride implementation therefore provides a viable pathway toward a deployable materials-discovery system for hydrogen-storage capacity screening. It reveals that the next goal for \scale{} is prospective closure: newly proposed candidates should be evaluated by higher-fidelity deterministic workflows and independent experimental measurements, and those results should be returned to the calibration and teacher-generation layers. This closed-loop validation will determine whether the architecture can move from accurate amortized screening to experimentally confirmed materials discovery.

\section{Conclusion}

\scale{} advances a general design principle for energy-materials intelligence: experimental calibration is not a terminal correction applied to a predictor, but the mechanism that converts deterministic operator results into an experimentally anchored teaching corpus. This preserves the provenance of each stage---scientific operator, measured evidence, calibration model, virtual-label construction, and deployed surrogate---while allowing them to work as a single discovery system. The resulting model is not asked to replace physics or experiment; it is asked to make their calibrated joint evidence reusable at the scale required for ranking and exploration.

The representative metal-hydride implementation establishes this sequence in executable form. A phase-equilibrium capacity operator and measured storage-capacity anchors generate a calibrated response surface; crystallographically anchored periodic graphs carry deployable composition, prototype, and operating-condition information into an edge-biased transformer; and the resulting surrogate reproduces the teacher distribution with high held-out fidelity. This demonstrates that a structured, experimentally calibrated scientific workflow can be compressed without discarding the chemical and thermodynamic context needed to audit its predictions.

More broadly, SCALE changes the allocation of scientific effort in large materials campaigns. High-cost computation and measurements can be concentrated on the points where they are most informative---operator construction, calibration, uncertainty reduction, and prospective validation---while a trained transformer performs broad candidate evaluation. The architecture is therefore a route to a closed, evidence-aware discovery loop: simulations and experiments improve the teacher system, the surrogate expands the searchable design space, and priority candidates return to scientific evaluation. This is the basis for scalable materials intelligence directed toward secure, affordable, resilient, and sustainable energy systems.

\section*{Acknowledgements}

Financial support for this work was provided by Alpha Ladder Group Pte Ltd. The authors thank Alpha Ladder Particle internal technical support and research discussions for practical feedback on the SCALE architecture and its energy-materials implementation. The authors also acknowledge internal computational resources used for model development and benchmarking.

\bibliographystyle{unsrtnat}
\bibliography{SCALE_MH_ArXiv_vF_09Sep2026}

\end{document}